%% file: main.tex
\documentclass{article}
\usepackage{spconf,amsmath,graphicx}
\usepackage{pgfplots}
\usepackage{enumitem}
\usepackage{multicol}
\usepackage[colorlinks=true,allcolors=blue]{hyperref}
\usepackage{url}
\usepackage{graphicx}
\usepackage{amsmath}
\usepackage{amssymb}
\usepackage{booktabs}
\usepackage{enumitem}
\usepackage{float}
\restylefloat{table}
\usepackage{stfloats}
\usepackage{graphicx}
\usepackage{bold-extra}
\usepackage{bm}
\usepackage{url}
\usepackage[utf8]{inputenc}
\usepackage{amsthm}
\usepackage{lipsum} 
\usepackage{booktabs}
\usepackage{algorithm}
\usepackage[utf8]{inputenc}
\usepackage{amssymb,color}
\usepackage{caption}
\usepackage{makecell,booktabs}
\usepackage{mathtools}
\usepackage{multicol}
\usepackage{amsmath, amssymb, dsfont}
\usepackage{mathabx}
\usepackage{epstopdf}
\usepackage{subcaption}
\usepackage{breqn}
\usepackage{verbatim}
\usepackage{arydshln}
\usepackage{tikz}
\usepackage{tikz,pgfplots}
\usepackage{pgfplots}
\usetikzlibrary{patterns}
\usepackage{algorithm}
\usepackage{algpseudocode}
\usepackage{amsfonts}

\name{Hao Chen$^\dagger$, Yusen Wu$^\ast$, Phuong Nguyen$^\ast$, Chao Liu$^\dagger$, Yelena Yesha$^\ast$}
\address{$^\ast$Dept. of Computer Science, University of Miami, Miami, FL, USA \\ 
$^\dagger$Dept. of Computer Science, University of Maryland, Baltimore County, MD, USA}

\input{mysymbols}

\begin{document}
\title{Soft Merging: A Flexible and Robust Soft Model Merging Approach for Enhanced Neural Network Performance}

\maketitle


\begin{abstract}
Stochastic Gradient Descent (SGD), a widely used optimization algorithm in deep learning, is often limited to converging to local optima due to the non-convex nature of the problem. Leveraging these local optima to improve model performance remains a challenging task. Given the inherent complexity of neural networks, the simple arithmetic averaging of the obtained local optima models in undesirable results. This paper proposes a {\em soft merging} method that facilitates rapid merging of multiple models, simplifies the merging of specific parts of neural networks, and enhances robustness against malicious models with extreme values. This is achieved by learning gate parameters through a surrogate of the $l_0$ norm using hard concrete distribution without modifying the model weights of the given local optima models. This merging process not only enhances the model performance by converging to a better local optimum, but also minimizes computational costs, offering an efficient and explicit learning process integrated with stochastic gradient descent. Thorough experiments underscore the effectiveness and superior performance of the merged neural networks.
$\mathsf{\href{https://github.com/AnonymousUser08/MergeNN}{https://github.com/AnonymousUser08/MergeNN}}$
\end{abstract}

\begin{keywords}
Model Merging,  Model Optimization   
\end{keywords}

\input{sections/intros}

\input{sections/method}
\input{sections/experiment}

\input{sections/conclusions}

\bibliographystyle{IEEEbib.bst}
\bibliography{ref}

\end{document}

%% file: mysymbols.tex
\newcounter{lemma} 

\newcounter{corollary} 

\newcounter{proposition} 

\newcounter{remarkjs} 

\def\ccal0{{\ensuremath{\mathcal 0}}}
\def\bbX{{\ensuremath{\mathbf X}}}
\def\bbY{{\ensuremath{\mathbf Y}}}

\def\bbg{{\ensuremath{\mathbf g}}}

\def\bbs{{\ensuremath{\mathbf s}}}

\def\bb0{{\ensuremath{\mathbf 0}}}
\def\bbalpha{{\mbox{\boldmath $\alpha$}}}
\def\bbbeta{{\mbox{\boldmath $\beta$}}}

\def\bbtheta{{\mbox{\boldmath $\theta$}}}



%% file: sections/intros.tex
\section{Introduction}
In the recent decade, deep learning has been flourishing in various domains. However, the inherent complexity of neural networks, with their intricate non-linearity and non-convexity, poses formidable challenges. The stochastic gradient descent (SGD) algorithm, despite using identical training data and network architectures, converges to distinct local optima due to different initializations. This leads to a fundamental question: \textit{Can the diverse local optima be leveraged to merge models, enhancing performance and moving closer to a more favorable global optimum?}

Convolutional neural networks exhibit various architectural paradigms like ShuffleNet \cite{ma2018shufflenet}, ResNet \cite{he2016deep}, UNet \cite{ronneberger2015u} and DenseNet \cite{huang2017densely}, each with unique features. Our primary challenge lies in devising an algorithm that accommodates these disparate designs. The secondary challenge involves ensuring the robustness of the merging algorithm across models with vastly varying parameter values. Additionally, we face the third challenge of selectively merging specific components rather than all parameters, aiming for efficiency.

Model merging, as a novel and challenging research direction, has seen limited exploration in existing literature. Unlike model combination and aggregation \cite{opitz1999popular,he2021automl,papernot2016semi}, which fuses different architectures, model merging improves the model performance by integrating the trained ones (local optima) with congruent architectures. Simple techniques like arithmetic averaging fall short due to the intricate nature of neural networks as addressed in the paper \cite{samuel2022git}. Further, it proposes to merge the models by solving a permutation problem, assuming that local optima exhibit similar performances, they match the neurons of two models. The \cite{pavsen2022merging} proposed a general framework for merging the models, with ``teacher" and ``student" concepts. Primarily the existing methods focus on neuron-level merging, which means they are targeting the weights of the neural network. However, relying solely on this approach has limitations in applicability, flexibility, and robustness, particularly with irregular models.

To address these issues, we introduce a novel paradigm called \textit{soft merging}, known for efficiency, adaptability, and robustness. Our method draws from model merging and channel pruning research \cite{louizos2017learning, pavsen2022merging, molchanov2019importance, voita2019analyzing}. It involves concurrent training of gate parameters for multiple models, using a differentiable surrogate of $l_0$ regularization to identify crucial parts. Instead of updating the weights, it only picks the best ones from the provided set of weights. This enables selective merging across various layers and architectures, with enhanced adaptability. In summary, our contributions include:
\begin{itemize}[itemsep=-1pt]
    \item Our proposal outlines a general procedure for selectively soft merging multiple models simultaneously with diverse neural network architectures.
    \item We present an algorithm that achieves linear complexity for efficient soft merging.
    \item  Extending neural network model merging to accommodate a wide range of deep learning designs ensures robustness, even in the presence of anomalies.
\end{itemize}

%% file: sections/method.tex
\section{Proposed Methods}
\subsection{Problem statement}
Suppose there are $J$ given models denoted as $\{\mathcal{M}_j\}_{j=1}^J$ with the same neural network architecture. Given training data $\bbX$ and labels $\bbY$, our goal to find the optimal model $\mathcal{M}^*$ from the object function
\begin{align}
    \min_{\{g_j\}_j} \sum_j \mathcal{L}(g_j\mathcal{M}_j(\bbX), \bbY; \bbtheta_j),  s.t. \sum_j g_j = 1, \label{model_level}
\end{align}
where $g_j \in \{0,1\}$ is a gate parameter, $\mathcal{L}$ is the loss function, and $\bbtheta_j$ represents the neural networks parameters in the $j$-th model. Labels $\bbY$ may not be necessary in some learning tasks, and they can be ignored in specific learning objective functions. The loss function $\mathcal{L}$ is a general function, which can be utilized in various machine learning methods, including supervised, unsupervised, and semi-supervised approaches. The Eq.(\ref{model_level}) is called the model-level merging, if $\bbtheta_j$ is fixed, which is equivalent to picking the best model among the $J$ models. Jointly learning $\bbtheta_j$ and $g_j$ belongs to the wide-sense {\em hard} merging, because $\bbtheta_j$ may change during the merging process. While, if learning gates parameter $g_j$ only, with $\bbtheta_j$ held fixed, it is referred to as {\em soft merging}. Especially, here Eq.(\ref{model_level}) performs the soft merging on the model, namely picking the best granule from all the granules, with each model as a granule, which is a high-level merging. In the following section, we will introduce soft merging at different levels.

\subsection{Model Merging at Various Levels of Granularity}
To merge the full model of the neural networks, we can also apply the merging process at a lower level by merging the individual modules or layers. Suppose the model $\mathcal{M}_j$ consists of $L$ layers; we can disassemble $\mathcal{M}_j$ into individual layers $\mathcal{M}_j:= \mathcal{F}({\{\Lambda}_{l,j}\}_{l=1}^{L})$, where $\mathcal{F}(\cdot)$ is a structural function to connect each layer which bears the same design for all $J$ models, and $\Lambda_{l,j}$ is the $l$th-layer in the model $\mathcal{M}_j$. Some of the layers in the $j$-th model could aggregate to a module, which is defined as $\Phi_{m,j}:= \mathcal{F}_m({\{\Lambda}_{l,j}\}_{l=m}^{m'})$ as the $m$-th module. Here, whether we are referring to the model $\mathcal{M}$, the module $\Phi$, or the layer $\Lambda$, they all share the fundamental characteristic of being composed of linear or non-linear functions. So the model $\mathcal{M}_j$ can be written as
\begin{align}
    &\mathcal{M}_j =  \mathcal{F}_M(\{\Phi_{m,j}\}_{m=1}^{M}) 
    =  \mathcal{F}({\{\Lambda}_{l,j}\}_{l=1}^{L}),
\end{align}
where $\mathcal{F}_M(\cdot)$ is the structural function to connect all the modules in the model $\mathcal{M}_j$. The objective of module-level merging is to address the following problem
\begin{align}
&\min_{\{ g_{m,j}\}_{m,j}}  \mathcal{L}( \mathcal{M}(\bbX), \bbY   ; \bbtheta)\ \ \  s.t. \sum_j g_{m,j} = 1 \nonumber \\
&\Phi_m = \sum_j g_{m,j} \Phi_{m,j},\ 
\mathcal{M} = \mathcal{F}_M(\{\Phi_m\}_{m=1}^{M}) 
\label{module_level}
\end{align}

where $g_{m,j}$ are the module-level gates. The gates are applied after the data through the module $\Phi_m$ and the whole flow is managed by $\mathcal{F}_M$. Similarly, the layer-level problem can be formulated as
\begin{align}
&\min_{\{g_{l,j}\}_{l,j}} \ \mathcal{L}(  \mathcal{M}(\bbX), \bbY   ; \bbtheta), \ \ s.t. \sum_j g_{l,j} = 1, \nonumber \\
&\textit{with }\Lambda_l = \sum_j g_{l,j} \Lambda_{l,j}, \  \mathcal{M} = \mathcal{F}(\{\Lambda_l\}_{l=1}^{L})
\label{layer_level}
\end{align}

\vspace*{-5mm}
\subsection{Merging Optimization Algorithms} 
We consider $g_j$ as a random variable following a Bernoulli distribution, however, it is not differentiable. To solve this problem, the constraint can then be reformulated in relation to the set of variables $\{g_j\}_j=\bbg \in \{0,1\}^J $, and $ \|\bbg\|_0=1$ indicating that only one element in $\bbg$ is allowed to be non-zero. Similarly, in Eq.(\ref{module_level}) and (\ref{layer_level}), the constraint can be restated in terms of $L_0$ norm. However, the $L_0$ norm is not differentiable nor convex. A typical surrogate of $L_0$ norms as $L_1$ norm, as a convex constraint, but imposing the sparsity would introduce another constraint. In the paper by Louizos et al. \cite{louizos2017learning}, a surrogate approach was introduced. This approach utilizes a random variable governed by a hard concrete distribution to address the $L_0$ norm constraint. Notably, this surrogate method retains differentiability through the implementation of the reparameterization trick. \\

\begin{figure}
\centering
\subfloat[$\log \alpha=-3$]{\label{fig:BLE_train}\includegraphics[width=0.49\linewidth ]{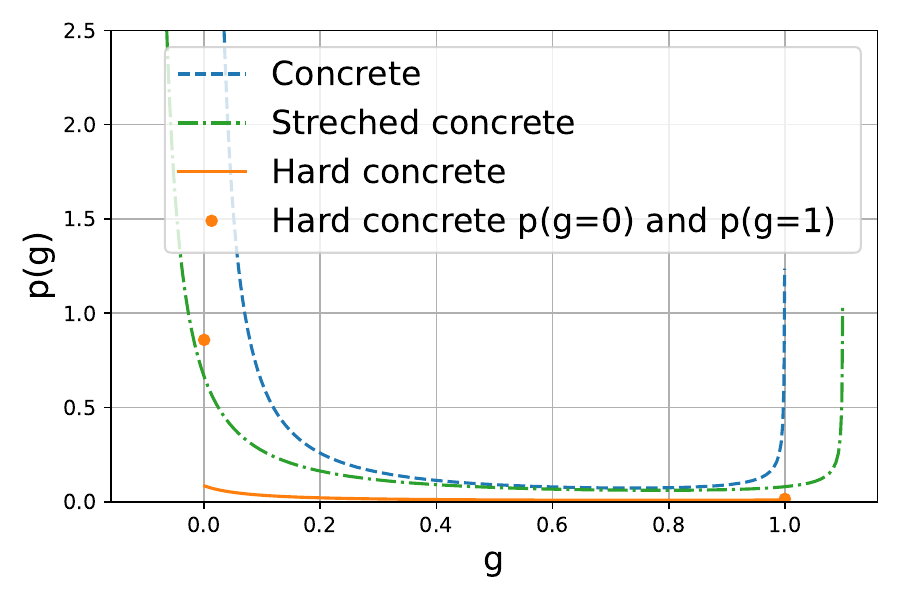}}
\subfloat[$\log \alpha=0$]{\label{fig:BT_train}\includegraphics[width=0.49\linewidth]{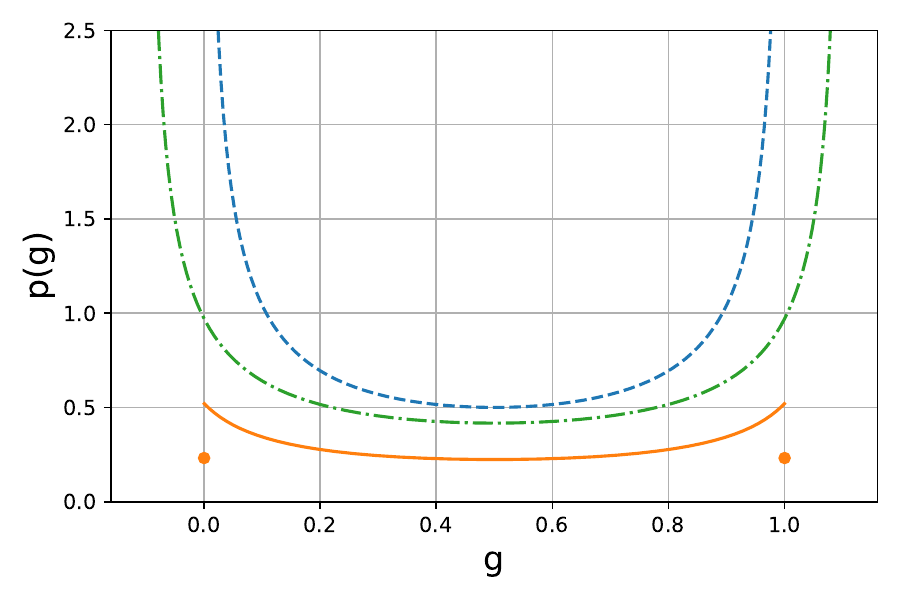}}   
\caption{The PDF of concrete, stretched concrete, and hard concrete distribution with $\beta = 0.5$ and different $\log \alpha$ values. (a) Here we have $\log \alpha = -3$, and the hard concrete distribution has $p(g=0)=0.8583$ and $p(g=1)=0.0148$. (b) For $\log \alpha=0$, we have $p(g=0)=p(g=1)=0.2317$ for the hard concrete distribution. }
\label{fig:pdf}
\vspace{-5mm}
\end{figure}

\noindent \textbf{Hard Concrete Distribution}.
The probability density function (PDF) of concrete distribution is written as,
\begin{align}
    p(s; \beta, \alpha) = \frac{\alpha \beta s^{\beta -1}(1-s)^{\beta -1}}{(s^\beta +\alpha(1-s)^\beta)^2}, \ \ 0<s<1. \label{eq:pdf_concrete}
\end{align}
with the cumulative distribution function (CDF) as
\begin{align}
    F(s; \beta, \alpha) = \frac{1}{e^{\log \alpha + \beta (\log(1-s) - \log s)}} \label{eq:cdf_concrete}
\end{align}
where $\alpha > 0$ and $0< \beta <1$. The parameter $\alpha$ controls the distribution 
This binary-like concrete distribution is a smooth approximation of Bernoulli distribution\cite{maddison2016concrete}, because it can be reparameterized with uniform distribution $u \sim \mathcal{U}(0,1)$ as
$s = \text{Sigmoid}(\log (u) - \log(1-u) + \log \alpha)$, where $\text{Sigmoid}(x) = \frac{1}{1+e^{-x}}$. However, the concrete distribution does not include $0$ and $1$. To tackle this problem, the \cite{louizos2017learning} proposed a method stretching $s$ to $(\gamma, \zeta)$ by $\bar s = s\zeta+ (1-s)\gamma$, with $\gamma<0$ and $\zeta>1$. Then by folding $\bar s$ into $(0, 1)$ by $g = \min(1, \max(\bar s, 0))$, the hard concrete distribution has the CDF simply as 
\begin{align}
    Q(s; \beta, \alpha)=F(\frac{s-\gamma}{\zeta-\gamma}), \ \ 0 \leq s \leq 1
\end{align} 
and the PDF as
\begin{align}
    &q(s;\beta, \alpha) = F(\frac{\gamma}{\gamma - \zeta})\delta(s)+\left(1-F(\frac{1-\gamma}{\zeta-\gamma})\right)\delta(s-1) \nonumber\\
    &+\left(F(\frac{1-\gamma}{\zeta-\gamma})-F(\frac{\gamma}{\gamma - \zeta})\right)p(\frac{s-\gamma}{\zeta-\gamma}), \ \ 0 \leq s \leq 1.
\end{align}
The comparisons among examples of concrete, stretched concrete, and hard concrete distribution are shown in Fig.\ref{fig:pdf}. \\

\noindent \textbf{Surrogate Loss Functions}
All the gates are replaced by the surrogate probability random variables following the hard concrete distribution. With the reparameterization trick \cite{kingma2013auto}, reformulated with the Lagrangian multiplier, the loss function for model-level merging is
\begin{align}
    \min_{\{\alpha_j, \beta_j\}} \sum_j \mathcal{L}(\hat s_j\mathcal{M}_j(\bbX), \bbY; \bbtheta_j) + \lambda ( \hat s_j -\frac{1}{J}) \nonumber \\
    \textit{with} \ \hat s_j \sim q(s_j>0;\alpha_j, \beta_j) \label{loss_model}
\end{align}
Similarly, we can get the module- and layer-level merging loss functions respectively as
\begin{align}
\min_{\{\alpha_{m,j}, \beta_{m,j}\}_{m,j}}  \mathcal{L}(  \mathcal{M} (\bbX), \bbY   ; \bbtheta) + \lambda \sum_{m,j}( \hat s_{m,j} -\frac{M}{J}) \nonumber \\
\textit{with} \ \hat s_{m,j} \sim q(s_{m,j}>0;\alpha_{m,j}, \beta_{m,j})
\end{align}
\begin{align}
\min_{\{\alpha_{l,j}, \beta_{l,j}\}_{l,j}}  \mathcal{L}(  \mathcal{M} (\bbX), \bbY   ; \bbtheta) + \lambda \sum_{l,j}( \hat s_{l,j} -\frac{L}{J}) \nonumber \\
\textit{with} \ \hat s_{l,j} \sim q(s_{l,j}>0;\alpha_{l,j}, \beta_{l,j}) \label{loss_layer}
\end{align}

\subsection{General training algorithm}
We have the full-model soft merging in different levels as shown in Eq.(\ref{loss_model}) - (\ref{loss_layer}). If one just selects a few layers or modules to perform the soft merging, the loss function should be changed accordingly. For example, supposing to merge the 1st and the 5th layers of the models, which means a layer-level merging, it requires the training parameter $\alpha_{1,j}, \alpha_{5,j}, \beta_{1,j}$ and $\beta_{5,j}$ as the random variable and others as fixed 
with given gate value as $1$, using formulation (\ref{loss_layer}). Here we propose the general problem formulation for full-model and selective soft merging in different levels as
\begin{align}
    \min_{\bbalpha, \bbbeta} \mathcal{L}_1 (\bbX, \bbY)+ \lambda \mathcal{L}_2(\bbalpha, \bbbeta) \label{loss_gen}
\end{align}
where $\mathcal{L}_1$ is related to model performance and $\mathcal{L}_2$ is the term controlling the merging, including the sampling process for the reparameterization. In the formulation (\ref{loss_gen}), the parameters $\bbalpha$ and $\bbbeta$ represent two sets of parameters selected for the process of selective soft merging. Notably, the hyper-parameter $\lambda$ remains fixed as a user-defined tuning parameter and is not learned during training. The training methodology is relatively straightforward, involving the application of SGD in the mini-batch fashion as outlined in Table \ref{tab:alg}.

\begin{table}[t]
\caption{General training algorithm.}
\centering 
\footnotesize
\begin{tabular}{l}
\hline
Input: $\bbX$, $\bbY$, $\{\mathcal{M}_j\}$, $\lambda$, the learning rate $\eta$\\
Output: $\mathcal{M}^*$  \\
\hline
\phantom{0}1: Initialize $\bbalpha$ and $\bbbeta$ randomly\\
\phantom{0}2: For $b=0,1,\ldots$ \ \ /* $b$-th mini-batch */\\
\phantom{0}3: \quad $\mathcal{L}_l = \mathcal{L}_1 (\mathcal{M}(\bbX^{(b)}), \bbY^{(b)})+ \lambda \mathcal{L}_2(\bbalpha, \bbbeta)$ /*$\bbX^{(b)}, \bbY^{(b)}$ \\
\quad \quad \quad as the data and labels for current mini-batch, $\mathcal{L}_l$ is the loss \\
\quad \quad \quad  function for full-model or selective merging*/\\

\phantom{0}4: \quad $\bbalpha = \bbalpha + \eta \frac{\partial \mathcal{L}_l}{\partial \bbalpha}, \bbbeta = \bbbeta + \eta \frac{\partial \mathcal{L}_l}{\partial \bbbeta}$ \\

\phantom{0}5: Next $b$\\
\phantom{0}6: $\mathcal{M}^*$ contains the gate parameters $\ \hat \bbs^* \sim q(\bbs;\bbalpha^*, \bbbeta^*)$ \\
\hline
\end{tabular}
\label{tab:alg}
\vspace{-2mm}
\end{table}

%% file: sections/experiment.tex
\section{Experiments} \label{experiment}
We conducted multiple experiments at various levels of merging to demonstrate the performance of multi-model soft merging, assess the robustness of the merging process, and explore selective merging across diverse neural networks. Nevertheless, the tasks of the experiments include supervised classification and unsupervised source separation. We used the ESC-$50$ \cite{piczak2015dataset}, and MNIST (with miture) as the data sets. The neural networks are Audio Spectrogram Transformer (AST) \cite{gong21b_interspeech}, ResNet18 \cite{he2016deep}, Variational auto-encoder (VAE) \cite{neri2021unsupervised}. We apply soft merging at various levels to evaluate the performance of our proposed algorithm. By experimenting with different settings, we aim to demonstrate the versatility of our soft merging approach across a broad spectrum of tasks.

\begin{figure}
  \centering
  \begin{tikzpicture}
    \begin{axis}[
      ybar,
      width=0.8\columnwidth, 
      height=4.5cm,
      bar width=0.25cm,
      xlabel={Models},
      ylabel={Accuracy (\%)}, 
      ymin=0,
      ymax=110,
      symbolic x coords={1, 2, 3, 4, 5, 6, 7, 8, 9, 10, Merged},
      xtick=data,
      xticklabel style={rotate=45, anchor=east}, 
      nodes near coords,
      nodes near coords align={vertical},
      every node near coord/.append style={font=\tiny}, 
      ]
      \addplot coordinates {
        (1, 16.1)
        (2, 17.4)
        (3, 52.3)
        (4, 53.8)
        (5, 55.8)
        (6, 84.4)
        (7, 84.6)
        (8, 85.1)
        (9, 94.1)
        (10, 95.8)
        (Merged, 95.8)
      };
    \end{axis}
  \end{tikzpicture}
  \vspace{-4mm}
  \caption{Accuracy Comparison of Models and Merged Model}
  \label{fig:accuracy}
\end{figure}
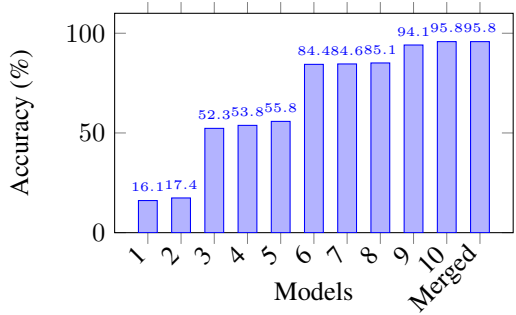

\noindent \textbf{Model-Level: 10 Models Merging}.
The proposed algorithm involves model-level soft merging of 10 vision transformer (ViT) models for audio source \cite{gong21b_interspeech} classification using the ESC-50 dataset. This approach employs parallel selection of the best model post-training, in contrast to the sequential comparison of neural network models. The dataset comprises 50 environmental sound classes, each containing 40 examples, which are divided into 1600 training and 400 validation samples. These models, initially pre-trained on ImageNet, process audio spectrograms using non-overlapping patches. Within the pool of 10 models, ranging from notably underperforming to highly competent ones, the soft-merging technique demonstrates its effectiveness even with limited training data. Furthermore, learning from validation data is accomplished within a mere 5 epochs, thereby reducing computational complexity compared to traditional sequential inference methods. The performance of the merged model, as illustrated in Fig.\ref{fig:accuracy}, showcases its capabilities, while the learned attention parameters $\log \alpha$ in Table \ref{tab:attention} provide insights into model quality. Remarkably, even with unfavorable initializations, the training process successfully identifies correct gradient directions. Model 10 emerges as the top-performing choice, demonstrating the algorithm's effectiveness in model selection without the necessity for extensive hyperparameter tuning.

\begin{table}
\begin{center}
\scriptsize
\caption{Model-level merging $\log \alpha$ values} \label{tab:attention}
\begin{tabular}{ c|c|c|c|c|c} 
 \hline
 \textbf{Model} \#&  $1$ & $2$ & $3$ & $4$ &$5$  \\ \hline
 \textbf{Init.}  & $0.0089$  & $-0.0185$ & $-0.0075$ &$0.0276$   &$\textbf{0.0059}$ \\
 \textbf{Final} & $-0.7604$ & $ -0.7877$ & $0.6155$ & $0.6509$ &$0.6722$ \\
 \hline
 \hline
  \textbf{Model} \#&  $6$ & $7$ & $8$ & $9$ & $\textbf{10}$     \\ \hline
 \textbf{Init.}& $0.0023$  & $0.0120$  & $-0.0098$  &$0.0115$   &$-0.0003$\\
 \textbf{Final}& $0.7943$ & $0.7943$ & $0.7878$ & $0.8740$ & $\textbf{0.8870}$\\
 \hline
\end{tabular}
\end{center}
\vspace{-5mm}
\end{table}

\begin{figure}[t]
\centering
\subfloat[Learning curves]{\label{fig:lr6module}\includegraphics[width=0.45\linewidth ]{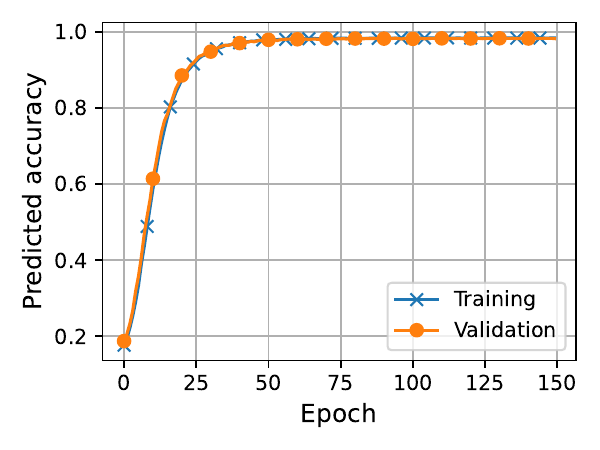}}
\subfloat[$\log \bbalpha$]{\label{fig:logalpha}\includegraphics[width=0.45\linewidth ]{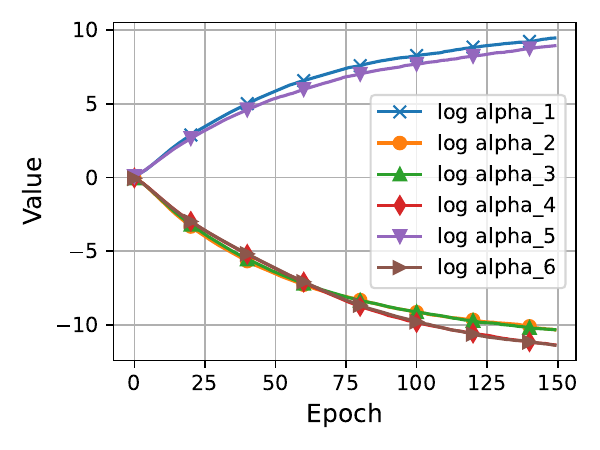}}   
\vspace{-3mm}
\caption{Module-level merging result, using three ResNet18 models and manually split into $6$ modules, with only two correct ones}
\label{fig:tr}
\end{figure}

\noindent\textbf{Module-Level: Three Models Merging} This experiment aims to determine if our algorithm could effectively identify correct modules within a collection of both trained and untrained modules. Specifically, we took one trained ResNet18 model with MNIST and two untrained ResNet18 models, splitting each into two halves to create a total of six modules. Among these, two modules held functional values while the other four contained random values. Our objective was to discern the functional modules using learned gate values. Despite the initial poor performance of the three individual models due to the untrained modules, applying soft merging yielded promising outcomes, indicating successful learning of the correct gates. During this experiment, we utilized parameters $\lambda=5$, with a learning rate of 0.001 across 150 epochs. The initial $\log \bbalpha$ values followed a Gaussian distribution with $\mathcal{N}(0, 0.01)$. The learning curve in Fig.\ref{fig:lr6module} depicted the merged model's progression, demonstrating that while the initial performance was subpar due to random gate initialization (Fig.\ref{fig:logalpha}), both training and validation accuracy improved significantly and quickly converged after around 80 epochs. Notably, the convergence of $\log \bbalpha$ values in Fig.\ref{fig:logalpha} did not occur within 150 epochs, indicating that the parameter does not possess inherent bounds due to the formulation in Eq. (\ref{eq:pdf_concrete}) and (\ref{eq:cdf_concrete}). 

\begin{figure}
    \centering
    \subfloat[Gates values: the first model as the prime model]{\label{fig:s1}\includegraphics[width=0.9\linewidth ]{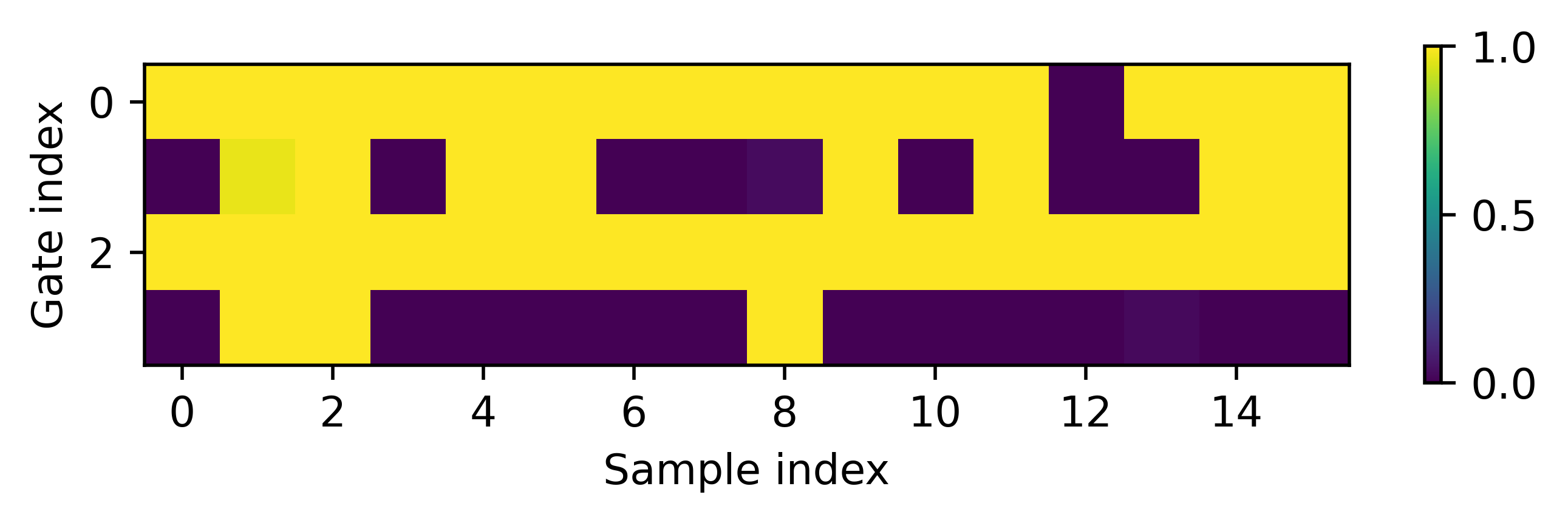}}
    
    \subfloat[Gates values: the other model as the prime model]{\label{fig:s2}\includegraphics[width=0.9\linewidth ]{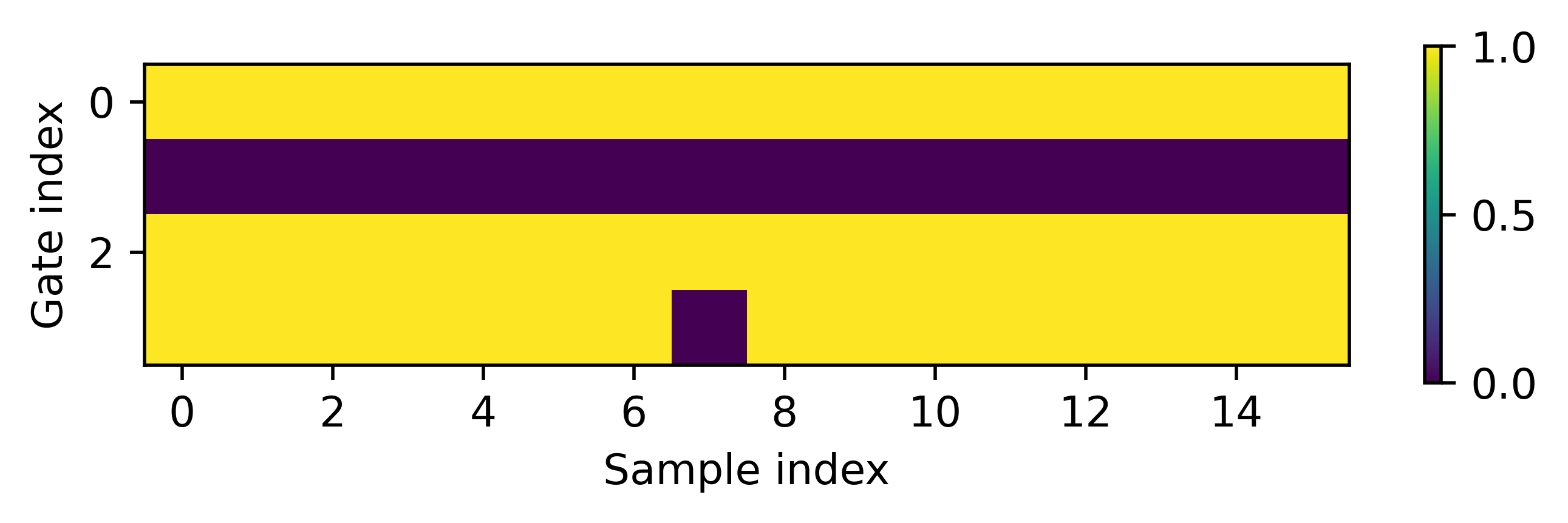}}

    \caption{The gate values of the last mini-batch training data.}
    \label{fig:layer}
\end{figure}

\noindent\textbf{Selective Layer-Level Merging }
In our unsupervised source separation experiment, we adapted Variational Autoencoders (VAEs) for blind source separation, showcasing the capabilities of our algorithm in such settings. We applied this concept to image data, similar to audio and RF blind source separation problems\cite{chen22unsupervised}. By manually creating MNIST mixtures without labels, we mirrored the approach in \cite{neri2021unsupervised}. We used two trained models with similar signal-to-interference ratio (SIR) performance and chose one layer in the encoder and one layer in the decoder to conduct the soft merging, which requires choosing a primary and secondary model. The VAE KL penalty $\beta_{KL}$ increased up to 0.5 per epoch for 10 epochs, with $\bbbeta = 0$ and $\lambda = 1$. The gate values in the last batch are depicted in Fig. \ref{fig:layer}, where different prime model selections led to varying $\log \bbalpha$, and still maintaining the SIR around $29$ but better than the one before merged.

%% file: sections/conclusions.tex
\section{Conclusions} \label{conclusion}
 Our research introduces the innovative concept of soft merging, a paradigm that addresses adaptability, efficiency, and robustness challenges in enhancing deep learning models. Our approach provides a versatile method for selectively integrating diverse neural network architectures, ultimately leading to improved model performance and a more favorable achievement of a better local optimum.